\documentclass[fleqn,10pt]{wlscirep}
\usepackage[utf8]{inputenc}
\usepackage[T1]{fontenc}
\usepackage{lineno}
\usepackage{multirow}
\usepackage{tabularx}
\newcommand{\figref}[1]{\figurename~\ref{#1}}
\newcommand{\tabref}[1]{\tablename~\ref{#1}}
\usepackage{xspace}
\newcommand{\tilburgscgk}{ETZ SC-GK\xspace}
\newcommand{\londonscgk}{LDN SC-GK\xspace}
\newcommand{\kchmcrc}{UK MC-RC-2\xspace}
\newcommand{\uclhmcrc}{UK MC-RC\xspace}

\title{Longitudinal Vestibular Schwannoma Dataset with Consensus-based Human-in-the-loop Annotations}

\author[1,*]{Navodini Wijethilake}
\author[1]{Marina Ivory} 
\author[2]{Oscar MacCormac} 
\author[3]{Siddhant Kumar} 
\author[1]{Aaron Kujawa} 
\author[1]{Lorena Garcia-Foncillas Macias} 
\author[2]{Rebecca Burger} 
\author[2]{Amanda Hitchings}  
\author[2]{Suki Thomson} 
\author[2]{Sinan Barazi} 
\author[2]{Eleni Maratos} 
\author[2]{Rupert Obholzer}  
\author[2]{Dan Jiang}  
\author[2]{Fiona McClenaghan}  
\author[2]{Kazumi Chia}  
\author[2]{Omar Al-Salihi} 
\author[2]{Nick Thomas} 
\author[2]{Steve Connor}  
\author[1]{Tom Vercauteren}  
\author[1,2]{Jonathan Shapey} 
\affil[1]{School of Biomedical Engineering and Imaging Sciences, King's College London, London, United Kingdom}
\affil[2]{King's College Hospital NHS Foundation Trust, London, United Kingdom}
\affil[3]{The Walton Centre NHS Foundation Trust, Liverpool, United Kingdom}
\affil[*]{corresponding author(s): Navodini Wijethilake (navodini.wijethilake@kcl.ac.uk)}

\begin{abstract}
Accurate segmentation of vestibular schwannoma (VS) on Magnetic Resonance Imaging (MRI) is essential for patient management but often requires time-intensive manual annotations by experts. While recent advances in deep learning (DL) have facilitated automated segmentation, challenges remain in achieving robust performance across diverse datasets and complex clinical cases. 
We present an annotated dataset stemming from a bootstrapped DL-based framework for iterative segmentation and quality refinement of VS in MRI.
We combine data from multiple centres and rely on expert consensus for trustworthiness of the annotations. Our human-in-the-loop framework integrates three primary components:
(1) a 3D nnUNet model for automated VS segmentation, trained with bootstrapping to improve segmentation accuracy with each round; 
(2) a multi-round quality assessment process involving consensus-driven expert review and manual corrections; and 
(3) an expert-driven post-hoc validation process on diverse datasets to assess model generalisability and performance robustness.
We show that our approach enables effective and resource-efficient generalisation of automated segmentation models to a target data distribution. The framework achieved a significant improvement in segmentation accuracy with a Dice Similarity Coefficient (DSC) increase from 0.9125 to 0.9670 on our target internal validation dataset, while maintaining stable performance on representative external datasets. Expert evaluation on 143 scans further highlighted areas for model refinement, revealing nuanced cases where segmentation required expert intervention. The proposed approach is estimated to enhance efficiency by approximately 37.4\% compared to the conventional manual annotation process. Overall, our human-in-the-loop model training approach achieved high segmentation accuracy, highlighting its potential as a clinically adaptable and generalisable strategy for automated VS segmentation in diverse clinical settings. The dataset includes 190 patients, with tumour annotations available for 534 longitudinal contrast-enhanced T1-weighted (T1CE) scans from 184 patients, and non-annotated T2-weighted scans from 6 patients. This dataset is publicly accessible on The Cancer Imaging Archive (TCIA) (\url{https://doi.org/10.7937/bq0z-xa62}). 
\end{abstract}
\begin{document}

\flushbottom
\maketitle

\thispagestyle{empty}

\noindent

\section*{Background and summary}
Data annotation is a critical yet often tedious, time-consuming, and expensive task in the development of machine learning models. Large-scale trustworthy annotations, which provide the reference standard (such curated annotations are often simply considered as \emph{ground truth} for machine learning training purposes) for training, validating, and testing these models, are particularly challenging to obtain in the domain of medical imaging. In this context, the accuracy of the annotations and representativeness of the training data  directly impacts the performance and reliability of machine learning algorithms, especially for complex tasks like tumour detection and segmentation. 

Medical data annotation, such as delineating VS and related structures in MRI scan, is inherently complex due to variability in tumour size and anatomical differences among patients. For VS, volumetric measurement is recognised as a more accurate and sensitive method for determining the true size of the tumour and is particularly effective at detecting subtle growth \cite{balossier2023assessing}. Currently, a 20\% increase in tumour volume is the standard threshold used to define growth in VS \cite{mackeith2018comparison}. However, recent studies suggest that for very small tumours, higher limits should be employed for growth assessment, as inter-observer variability can exceed the standard criteria \cite{cornelissen2024defining}. Additionally, agreement limits within volumetric annotation are influenced by tumour volume. The inter-observer limits of agreement improve as the tumour volume increases, making the growth criteria more reliable for larger tumours \cite{cornelissen2024defining}. These complexities often lead to variability and subjectivity in manual annotations, which can affect the consistency and generalisability of machine learning models trained on this data.  
 
For an automated model to be useful in this context, it needs to perform well across a diverse range of MRI hardware and imaging sequences used to acquire routine surveillance MRI . However, existing annotated datasets and pre-trained models may not accurately generalise. A typical approach to bridge the performance gap is to fine-tune a deep learning model using sufficient annotated data from the target distribution.
Annotating routine surveillance VS MRI present significant challenges due to their complexity. For instance, some patient scans may contain multiple pathologies, such as VS coexisting with other tumours like meningioma. Identifying and accurately annotating these cases is crucial, as the presence of multiple pathologies can introduce confounding factors that affect model performance and the validity of the results. Thus, there is a need for an efficient and trustworthy annotation protocol that can lead to large-scale annotated datasets representative of the deployment conditions.

We have previously published a collection of MRI scans from patients with a solitary VS undergoing Gamma Knife Stereotactic Radiosurgery (SRS) at a single institution \cite{shapey2021segmentation}.
In this instance, reliable annotations could be captured from the contouring performed as part of the standard radiosurgery planning and treatment process.
Unfortunately, this presents an important domain gap when compared to the annotation of surveillance MRI. To bridge this gap, we previously used an iterative annotation pipeline to generate high-quality manual reference segmentations for a large multi-centre routine clinical (MC-RC) longitudinal VS dataset \cite{kujawa2024deep}.  However, as detailed later in this manuscript, the resulting segmentation models still present a bias towards the referring clinical site leading to a decrease in performance when applied to a new target domain. Moreover, the annotation process employed for the MC-RC dataset remained labour intensive and expensive.

\begin{figure}[tbh!]
\centering
\includegraphics[width=0.7\linewidth]{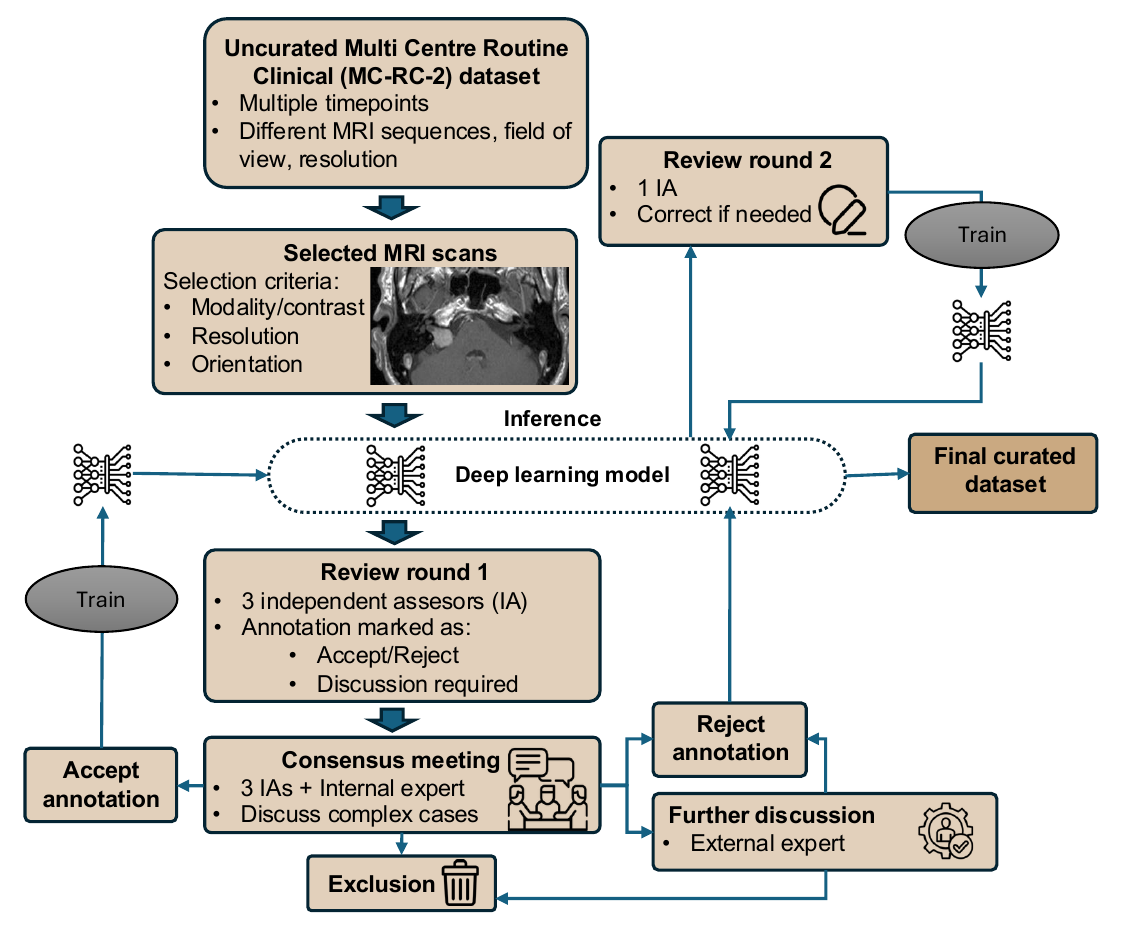}
\caption{Pipeline for data curation, DL-based collaborative iterative generation of quality-controlled VS segmentations.}
\label{fig:outline}
\end{figure}

In this work, we propose a human-in-the-loop approach to accelerate the generation of trustworthy annotations.
Relying on these already available annotated VS datasets, we implemented a consensus-based pipeline that integrates DL with expert validation to precisely and efficiently annotate our dataset. 
\figref{fig:outline} illustrates the pipeline, which consists of DL-based segmentation and multiple rounds of expert review to exclude or correct the segmentations as necessary.

\begin{figure}[tb!]
\centering
\includegraphics[width=0.7\linewidth]{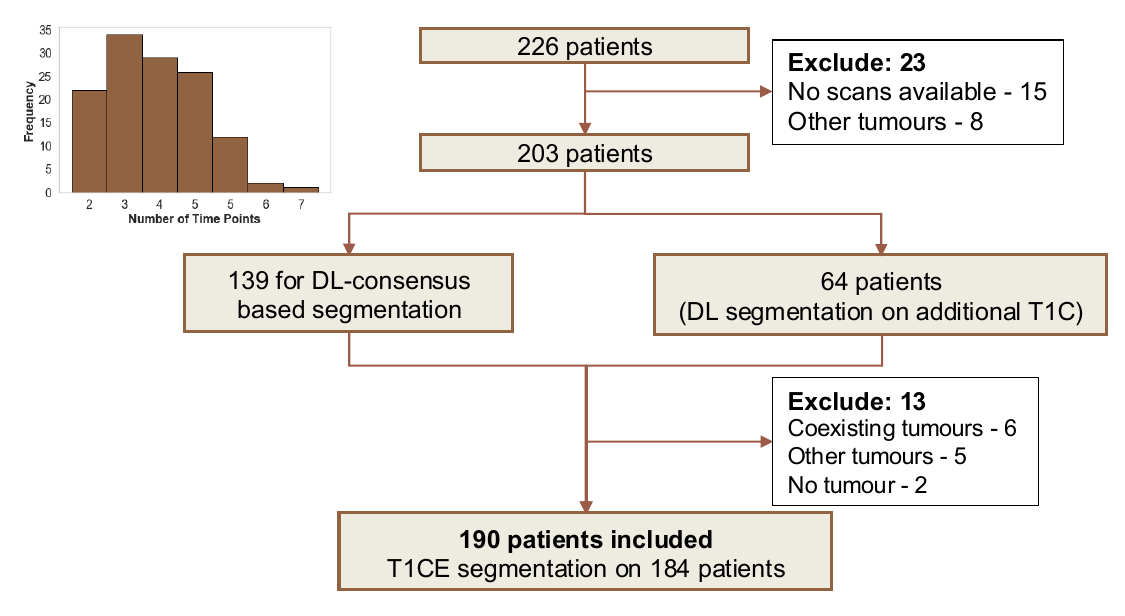}
\caption{Overall patient flow from the original uncurated \kchmcrc dataset to curated \kchmcrc dataset.}
\label{fig:patientflow}
\end{figure}

Future implementations may consider integrating this bootstrapping framework with real-time human-in-the-loop approaches to allow for efficient interactive correction and adaptation of segmentations in clinical workflows, facilitating adoption in routine neuro-radiology/surgery practice.

\section*{Methods}

We propose a human-in-the-loop data annotation methodology that integrates DL-based automated segmentation with expert-driven quality assurance processes. This approach is designed to enhance the efficiency and accuracy of manual annotation, particularly in the context of VS segmentation on MRI. The methodology involves several distinct phases, including model training, inference, and expert validation, as detailed below.

\subsection*{Ethics statement}
This study was approved by the NHS Health Research Authority and Research Ethics Committee (18/LO/0532). Because patients were selected retrospectively and the MR images were completely anonymised before analysis, no informed consent was required for the study.

\subsection*{Datasets}

\subsection*{External data}
\paragraph{London single centre gamma knife dataset (\londonscgk)}
The London SC-GK dataset was obtained from the Queen Square Radiosurgery Centre in London, United Kingdom, for radiosurgery planning purposes \cite{shapey2021segmentation}. Contrast-enhanced T1-weighted (T1CE) imaging was performed using an MP-RAGE sequence, featuring an in-plane resolution of 0.47×0.47 mm, a matrix size of 512×512, and a slice thickness between 1.0 and 1.5 mm. The imaging parameters included a repetition time (TR) of 1900 ms, an echo time (TE) of 2.97 ms, and an inversion time (TI) of 1100 ms. The scans were acquired on a Siemens Avanto 1.5T MRI scanner equipped with a 32-channel system and a Siemens single-channel head coil. This dataset is publicly avaiable on TCIA (\url{https://doi.org/10.7937/TCIA.9YTJ-5Q73}).

\paragraph{Tilburg single centre gamma knife dataset (\tilburgscgk)}
The Tilburg SC-GK dataset was acquired at TweeSteden Hospital, Tilburg, Netherlands, for radiosurgery treatment planning. T1CE imaging was performed using a 3D-FFE sequence, with an in-plane resolution of 0.8×0.8 mm, a matrix size of 256×256, and a slice thickness of 1.5 mm (TR = 25 ms, TE = 1.82 ms). All scans were obtained on a Philips Ingenia 1.5T MRI scanner equipped with a Philips quadrature head coil.

\paragraph{UK multi-centre routine clinical dataset (\uclhmcrc)}
The UK MC-RC dataset is a comprehensive collection of imaging data gathered from ten medical sites across the United Kingdom, covering a time span from February 2006 to September 2019 \cite{kujawa2024deep}. This dataset demonstrates substantial variability in slice thickness, voxel volume, and intensities among the T1CE. The scans were acquired using a diverse range of MRI systems, including SIEMENS, Philips, General Electric, and Hitachi scanners, with magnetic field strengths of 1.0T, 1.5T, and 3.0T. This dataset is publicly avaiable on TCIA (\url{https://doi.org/10.7937/HRZH-2N82})).

\subsection*{Internal data}
\subsubsection*{UK multi-centre routine clinical \#2 dataset (\kchmcrc) (\url{https://doi.org/10.7937/bq0z-xa62})}
The dataset comprises longitudinal MRI scans from patients with unilateral sporadic VS, collected from over 15 medical sites across South East England, United Kingdom. A total of 226 patients were referred to the skull base clinic at King's College Hospital, London, where they underwent initial management between August 2008 and November 2012. Eligible participants were adult patients, aged 18 years or older, with a single unilateral VS. This included patients with prior surgical or radiation treatment but individuals with Neurofibromatosis type 2 (NF2) related schwannomatosis were excluded from the study. \figref{fig:patientflow} visualises the overall patient flow from the initially acquired data to the final curated dataset. 

\subsubsection*{Image selection}
All patients with MRI scans available for at least one time point were included in the study. Scans showing other tumours and those covering non-brain regions (e.g., neck) were excluded. Additionally, images with a slice thickness greater than 3.5 mm were excluded due to reduced sensitivity to small lesions and the impact of partial volume effects, which hinder accurate delineation and volumetric analysis of VS. 

After this screening process, 203 patient cases remained eligible for tumour annotation. A total of 590 T1CE scans and 148 T2-weighted scans were retrieved. Given the higher availability of T1CE scans, these were prioritised for tumour annotation.
Of these, 139 patients with T1CE scans available at least two time points were selected for DL-based collaborative segmentation. The remaining 64 patient cases were curated at a later stage; 105 scans were randomly selected from this cohort and used for validating the DL-based collaborative segmentation model.

\subsection*{Automatic annotation proposal}

\begin{figure}[tb!]
\centering
\includegraphics[width=\linewidth]{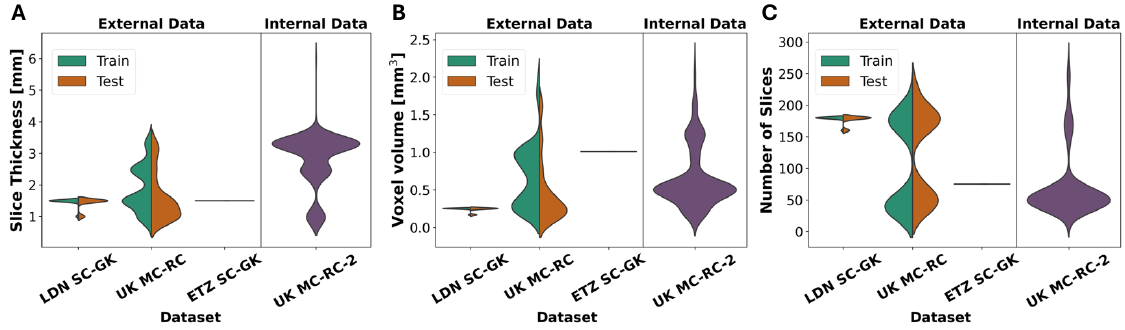}
\caption{Comparison of internal UK multi-centre routine clinical-2 (\kchmcrc) annotating dataset and external single-centre Gamma Knife (\londonscgk \& \tilburgscgk) and the \uclhmcrc datasets. Distributions of, \textbf{A)} slice thickness, \textbf{B)} image resolution in terms of voxel volume, and \textbf{C)} number of slices in each image. The standardised acquisition protocol of the SC-GK dataset results in homogeneous distributions across scans, while the heterogeneity in the MC-RC datasets reflects varied acquisition settings and protocols, leading to differences in slice thickness, resolution, and slice counts.}
\label{fig:datadistribution}
\end{figure}

\begin{table}[tb!]
\caption{Distribution of data between training, validation and testing sets across the three datasets used for building the segmentation model. \textbf{$N_{T}$}: Total number of scans. }\label{tab1:dataset_distribution}
\centering
\begin{tabular}{|l|l|l|l|l|}
\hline
Dataset	&	Train	&		Valid	&		Test	&	$N_{T}$ \\ \hline
\multicolumn{5}{|l|}{\textbf{Round 1 (External cohort)}}\\ \hline
\uclhmcrc	&	38	&	22	&	15	&	75\\ \hline
\tilburgscgk &	52	&	29	&	24	&	105\\ \hline
\londonscgk &	43	&	25	&	11	&	79\\ \hline
\textbf{$N_{T}$} &	133	&	76	&	50	&	259\\ \hline
\multicolumn{5}{|l|}{\textbf{Round 2 additions (Internal cohort)}}\\ \hline
\kchmcrc 	&	142	&	79	&	-	&	221\\ \hline
\multicolumn{5}{|l|}{\textbf{Round 3 additions (Internal cohort)}}\\ \hline
\kchmcrc 	&	235	&	132	&	-	&	367\\ \hline
\multicolumn{5}{|l|}{\textbf{Model evaluation (Internal cohort)}}\\ \hline
\kchmcrc 	&	&	&	-	&	105\\ \hline
\end{tabular}
\end{table}

\paragraph*{Training of the segmentation model}
We utilised the default 3D full-resolution UNet from the nnU-Net framework, along with the default pre-processing steps \cite{isensee2021nnu}. 
The initial training dataset, comprising expert-annotated images from the \uclhmcrc, \tilburgscgk, and \londonscgk external datasets, was used as the \emph{ground truth} for training the model.
With each round, the model was bootstrapped by incorporating additional cases from the target \kchmcrc internal dataset. 
For validation, two approaches were utilised to evaluate the performance of the bootstrapped model after each round:
\begin{enumerate}
\item 50 scans from the \uclhmcrc, \tilburgscgk, and \londonscgk datasets, referred to as the "Test set (external)." 
\item 105 previously unseen scans from the \kchmcrc dataset, which did not overlap with the patients used for bootstrapping, referred to as the "\kchmcrc validation set (internal)." 
\end{enumerate} 

\subsubsection*{Inference phase}
To generate the annotation proposals, a total of 427 T1CE scans from the \kchmcrc dataset were initially processed by the deep learning model trained solely with external data. 

\subsection*{Round 1: Quality assessment and the consensus meeting}
The annotation proposals obtained on the 427 scans were quality assessed by 3 independent experts; 2 neurosurgical fellows (S.K., O.M.) and a trained radiologist (M.I). The annotations were classified into three categories: Accept, Reject, and Other. 
'Accept' referred to cases with unilateral VS tumour and acceptable tumour annotation proposals.
'Reject' referred to cases with a unilateral VS tumour but without acceptable annotation proposals.
The 'Other' category included cases with other tumour types, VS tumours coexisting with additional tumours, instances where the tumour volume was cropped in the MRI scan or any session flagged for further discussion.
Annotation proposals marked as 'Accept' by all three experts were considered accepted. If even one expert marked a session as 'Reject,' it was classified as rejected. Sessions where at least one expert marked 'Other' were forwarded for consensus discussion. 

\begin{figure}[htb!]
\centering
\includegraphics[width=0.7\linewidth]{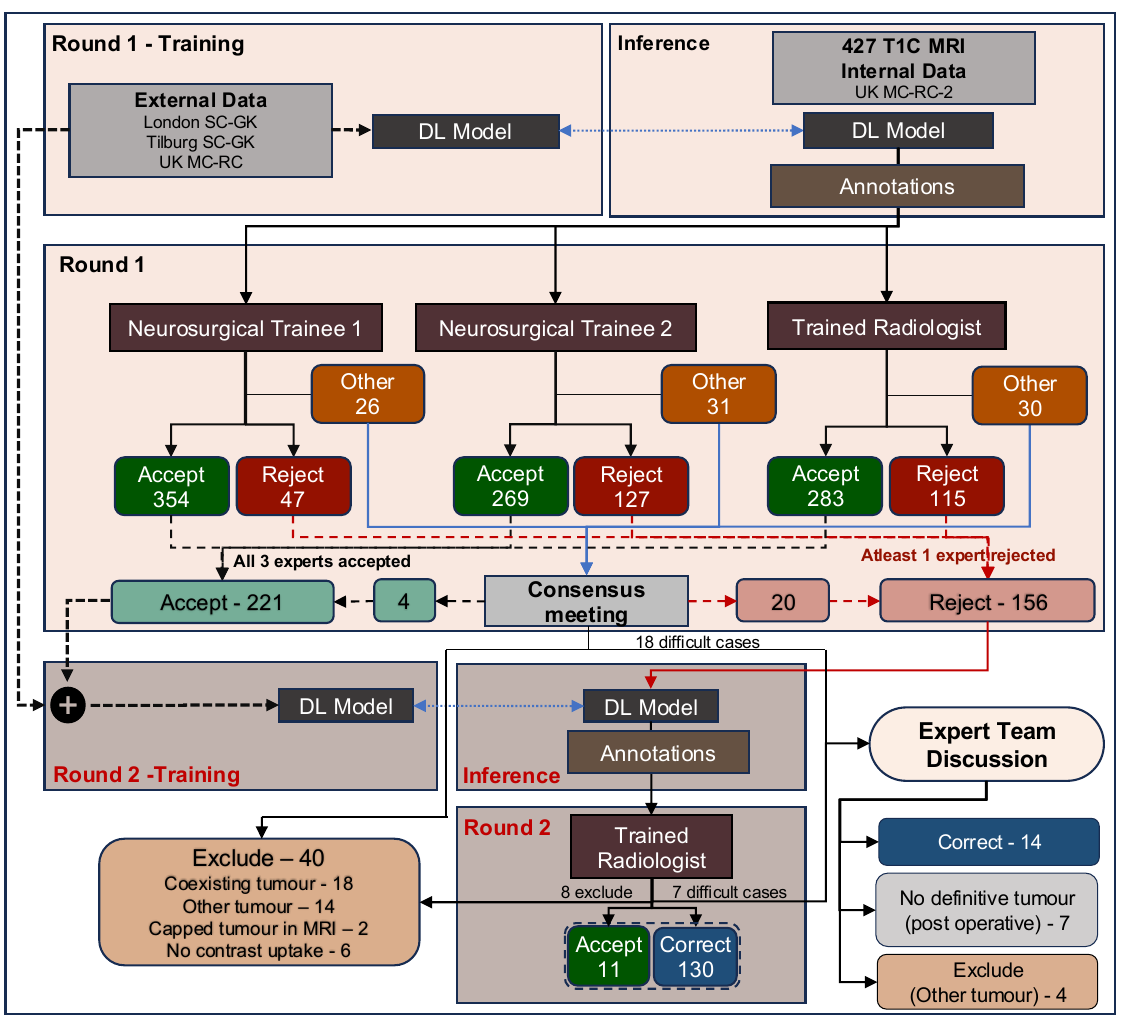}
\caption{Session flow from the data annotation pipeline.}
\label{fig:dataannotation}
\end{figure}

\subsubsection*{Consensus meeting}
The consensus meeting involved the three independent expert assessors who had previously reviewed the sessions, along with a consultant neurosurgeon (J.S.). The meeting focused on reviewing sessions marked as 'Other' by any of the experts. Each session was presented to the team, and the experts provided explanations for why the session required further discussion. Following this review, the sessions were re-categorised as 'Accept', 'Reject', or 'Exclude', with 'Exclude' indicating that the session or the entire patient case should be removed from the dataset. 
Following the consensus meeting, 221 cases were accepted, 156 were rejected, 32 were excluded and 18 difficult cases were forwarded for expert opinion as shown in \figref{fig:dataannotation}. 

\subsection*{Round 2: Bootstrapping and quality assessment}
After the consensus meeting, the \kchmcrc accepted annotations (221 scans) were combined with the initial training data from the three external datasets (\uclhmcrc, \tilburgscgk, and \londonscgk) to refine and enhance the segmentation model through bootstrapping (\tabref{tab1:dataset_distribution}).
The rejected sessions from Round 1 (156 scans) were then processed using this bootstrapped fine-tuned model to generate new annotation proposals.
An trained radiologist manually assessed the annotation proposals, marking them as accepted or correcting them using ITK-SNAP if needed. Since the cases underwent three independent reviews during round 1, no consensus was deemed necessary at this stage. Out of 156 cases, 11 were accepted, 130 required corrections, 7 difficult cases were forwarded for expert opinion, and 8 were excluded. Of the 8 exclusions, 6 were due to the absence of contrast uptake in T1 imaging, and 2 were caused by cropped tumour regions in the scans. Consensus was not required for these exclusions, as the issues were clear and unambiguous.

\subsection*{Round 3: Bootstrapping}
In the final round, the accepted and corrected annotations from Round 2, along with the accepted sessions from Round 1 (altogether 367 scans), were combined with the initial three external datasets (\uclhmcrc, \tilburgscgk, and \londonscgk) to further refine the model through bootstrapping. 
We conducted an expert team discussion, with an expert neuroradiologist of 23 years (S.C.), along with other team members involved in the annotation process, to review 25 complex scans, identified during the consensus meeting in Round 1 or by the trained radiologist in Round 2. After the expert neuroradiologist determined the tumour margins, trained radiologist corrected 14 scans. Seven scans were marked as having no definitive tumour, as they were post-operative cases. Four scans were excluded because they did not involve VS.

\subsection*{Finalised dataset}
The finalised dataset includes 190 patients, of whom 184 have 533 T1CE images with tumour annotations. Among the 139 patients who underwent the consensus-based annotation pipeline, tumour annotations were successfully generated for 125 patients. For 2 patients, no definitive tumours were identified, and 12 patients were excluded from the dataset. 

\paragraph{Demographic and clinical characteristics of the patient cohort}
The cohort included patients with a near-equal distribution of sex (82 males and 79 females), and 29 patients for whom sex information was not available. The mean age at the time of the first MRI was 58.1 years. The most frequently reported symptom was hearing loss, affecting 175 patients, followed by tinnitus (56 patients), balance disturbance (48 patients), facial numbness (25 patients), and facial weakness (23 patients). Less commonly reported symptoms included dizziness (14 patients), headache (15 patients), and speech difficulties (2 patients). Regarding treatment history, 14 patients underwent surgery alone, 37 patients received SRS alone, and 2 patients were treated with both surgery and SRS, indicating that the majority of patients received a single treatment modality. This dataset therefore provides a comprehensive overview of demographic and clinical characteristics, as well as treatment patterns, within this population.

\paragraph{Annotation of the additional scans} At a later stage, 64 patient cases were curated and added to the dataset. Among these, 39 patients had single time-point T1CE scans, while 21 patients had 2 to 5 longitudinal scans. Notably, 4 of the 64 cases included only T2 scans. Annotations for the additional 105 scans from these 64 patients were generated using the Round 3 bootstrapped model and subsequently reviewed and corrected by a trained radiologist (M.I.). As previously mentioned, this set is referred to as the "\kchmcrc validation set (internal)". 

\section*{Data records}
This proposed \kchmcrc dataset is hosted publicly on TCIA. The curated dataset comprises 1 to 8 longitudinal scans per patient acquired from 2010 onwards, totaling 543 contrast-enhanced T1-weighted (T1CE) scans,  481 T1-weighted scans and 133 T2-weighted scans across 621 time points (mean 3.25 scans per patient; mean monitoring period 4.83 $\pm$ 3.08 years). Scan dates are uniformly shifted for privacy, with consistent offsets applied within each patient’s imaging series. Imaging files are provided in NIfTI format (\texttt{.nii.gz}) and follow the naming convention \texttt{VS\_MC\_RC2\_\{studyID\}\_\{mri\_date\}\_\allowbreak\{scan\_type\}.nii.gz}, where \texttt{scan\_type} indicates T1, T1C, T2, or the corresponding tumor segmentation (\texttt{T1C\_seg}). Segmentation masks are provided for 534 T1CE scans, while masks are not included for 9 post-operative scans with no visible residual tumor. 

Demographic and clinical information for each MRI time point is included in \texttt{VS\_MC\_RC2\_demographics.xlsx} and contains sex, ethnicity, age at MRI, patient-reported clinical symptoms, treatment strategy (surveillance, stereotactic radiosurgery, or surgery) along with treatment dates and prior interventions, and post-surgical outcomes indicating presence or absence of residual tumor. Acquisition and scanner details for each imaging file are provided in \texttt{VS\_MC\_RC2\_Metadata.xlsx}, with one row per NIfTI file; metadata includes scanner manufacturer, magnetic field strength, protocol parameters, slice thickness, sequence-specific information, and patient positioning, with the \texttt{file\_name} column linking directly to the corresponding NIfTI file.

\section*{Technical validation}
\subsection*{Quantitative performance in each round of bootstrapping}
Performance on the test set (external) and the \kchmcrc validation set (internal) for the ensemble of 5 folds are shown in \tabref{tab:Dicescore-test}. 
For the internal validation set, the median DSC shows a notable improvement from Round 1 to Round 2, increasing from 0.9125\;(CI: [0.8075, 0.9483]) to 0.9639\;(CI: [0.8931, 0.9844]), with a slight further increase in Round 3 to 0.9670\;(CI: [0.8671, 0.9915]).
This progression indicates a consistent enhancement in model performance with each bootstrapping round, likely attributable to more robust training adjustments.
The Wilcoxon signed-rank test was performed between the rounds, revealing a statistically significant improvement (p $<$ 0.05) between Rounds 1 and 2, as well as between Rounds 1 and 3.

In contrast, the external test set’s DSC median scores exhibit a stable performance across rounds, with Round 1 starting at 0.9298\;(CI: [0.8958, 0.9491]) and slightly decreasing to 0.9265\;(CI: [0.8981, 0.9533]) in Round 2, followed by a minor drop to 0.9229\;(CI: [0.8947, 0.9531]) in Round 3.
These results suggest that while the proposed human-in-the-loop fine-tuning effectively generalises to our target internal data, the rounds of bootstrapping with internal cases did not significantly impact its performance on the external test set. 

\begin{figure}[tb!]
\centering
\includegraphics[width=0.8\linewidth]{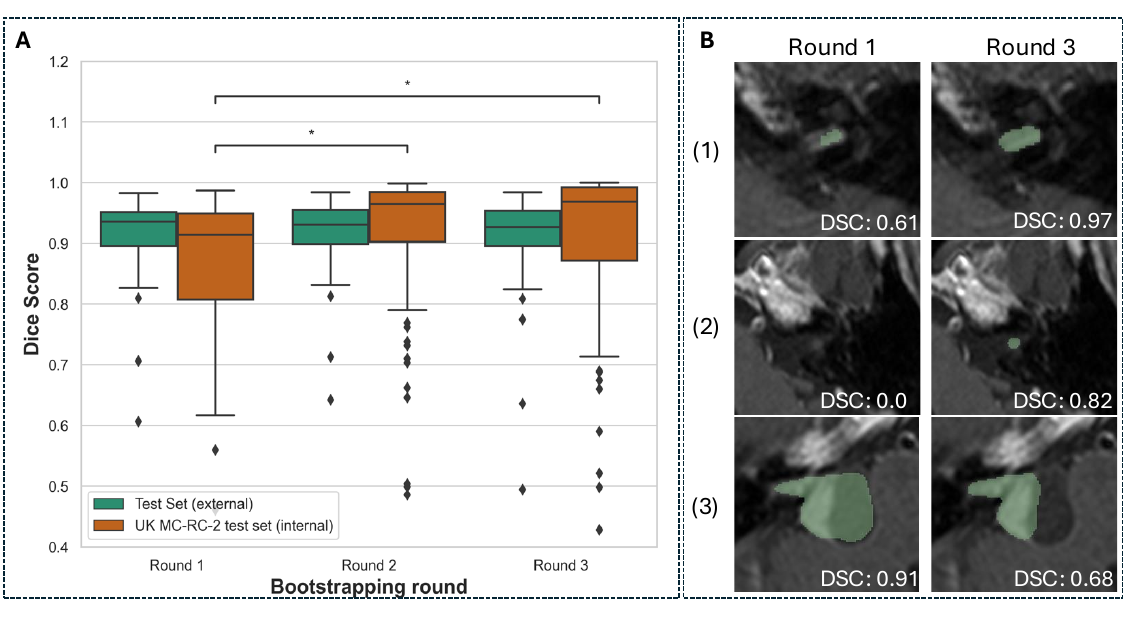}
\caption{\textbf{(A)}  DSC distribution on the external test set and the internal \kchmcrc validation set. A statistically significant improvement (p < 0.05) is observed between Rounds 1 and 2 and between Rounds 1 and 3.
\textbf{(B)} Three sample segmentations from the internal \kchmcrc validation set: (1) and (2) show improved DSC, while (3) demonstrates performance loss, the segmentation missing the large peritumoural cystic region.}
\label{fig:internalexternaldice}
\end{figure}

\begin{table}[tb!]
\centering
\caption{Median and interquartile of tumour DSC for each round}
\begin{tabular}{|l|l|c|} 
\hline 
\textbf{Validation Set} & \textbf{Round}                         & \textbf{DSC} \\ \hline
\multirow{3}{*}{\kchmcrc validation Set (internal)} &  Round 1 & 0.9125 [0.8075-0.9483]      \\\cline{2-3}
& Round 2 & 0.9639 [0.8931-0.9844]    \\\cline{2-3} 
& Round 3 & 0.9670 [0.8671-0.9915] \\ \hline
\multirow{3}{*}{Test Set (external)} & Round 1 & 0.9298 [0.8958-0.9491] \\ \cline{2-3} 
 & Round 2 & 0.9265 [0.8981-0.9533]   \\ \cline{2-3} 
& Round 3  & 0.9229 [0.8947-0.9531]   \\ \hline
\end{tabular} \label{tab:Dicescore-test}
\end{table}

The model demonstrated a statistically significant improvement in DSC from Round 1 to Round 2. However, subsequent improvements between Round 2 and Round 3 were not statistically significant, suggesting that the model's performance had likely reached a plateau. This plateau highlights a limitation in model refinement through additional data alone, even with corrected annotations. Given these results, further rounds of bootstrapping were deemed unnecessary, and the process was halted to prevent diminishing returns in computational resources and expert review time.

An improvement in the DSC may not capture the nuanced segmentation adjustments required for complex cases. This underscores the value of interactive correction, which enables experts to fine-tune segmentation masks on a case-by-case basis, addressing specific anatomical details that automated generalisation may overlook. Thus, although higher DSC can indicate progress, they may not always reflect the clinical relevance of segmentation precision. Interactive correction offers a pathway to enhance the model’s applicability in real-world clinical settings by allowing for more context-specific adjustments beyond generalised model outputs.

\subsection*{Efficiency of the proposed approach against the traditional manual annotation}
\begin{figure}[tbh!]
\centering
\includegraphics[width=\linewidth]{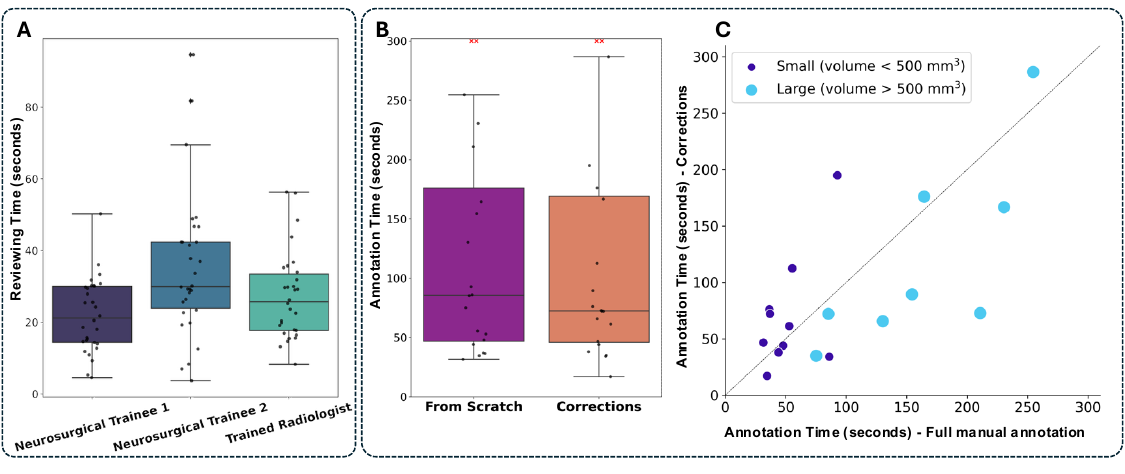}
\caption{\textbf{(A)} Distribution of time spent for quality assessment by three independent experts in round 1 \textbf{(B)} Distribution of time spent on full manual annotation and on correcting DL generated segmentations in round 2 of this proposed annotation pipeline by the trained radiologist. \textbf{(C)} Distribution of annotation times for full manual annotation versus manual correction of DL-generated segmentations across tumour sizes.}
\label{fig:efficiency}
\end{figure}

The resource efficiency of the proposed DL-based collaborative human-in-the-loop annotation pipeline was assessed in comparison to the traditional full manual annotation. 
In round 1, we measured the time taken for quality assessment by three independent experts including two neurosurgical fellows (S.K. and O.M.) and a trained radiologist (M.I.). Quality assessment of the DL-generated segmentations was performed on 30 randomly selected cases with the time taken to review recorded. Cases were randomly selected from a total of 427 cases. The three assessors—neurosurgical trainee 1, neurosurgical trainee 2, and the trained radiologist— spent a median time of 21.13 seconds, 29.95 seconds, and 25.70 seconds, respectively, per case for quality evaluation. \figref{fig:efficiency} (A) visualises the distribution of the time spent by 3 independent assessors. 

In round 2, 20 cases (selected from the 156 cases with rejected annotations from round 1, re-annotated using the updated DL model) were reviewed and corrected as necessary by the trained radiologist (M.I.), with time spent recorded. To provide a baseline comparison, the same 20 cases were also annotated fully manually by the trained radiologist (M.I.), with the time taken for tumour annotation on each scan recorded separately. To reduce potential bias, these two tasks (correction and full manual annotation) were performed 30 days apart. The median time for full manual annotation was 85.65 seconds, while the median time for correction of the DL segmentations was 72.40 seconds. Overall, full manual annotation took slightly more time, although the difference was not statistically significant. \figref{fig:efficiency} (B) shows the time trained radiologist spend on annotating 20 cases fully manually and correcting DL segmentations. \figref{fig:efficiency} (C) visualises the relationship between the two variables, where the 20 cases are  separated into 2 categories small and large, based on median tumour volume.

The proposed DL-based collaborative human-in-the-loop annotation approach demonstrates clear efficiency benefits over traditional manual annotation methods. The total time required for the proposed approach, which involved multiple experts, was 6.35 hours for 427 cases. In contrast, manual annotation of the same dataset would take approximately 10.15 hours, representing a 37.4\% reduction in time. Furthermore, a trend was observed toward reduced correction times compared to full manual annotation, particularly for larger tumours (\figref{fig:efficiency}(C)).




\subsection*{Expert radiological assessment on a 45 patient cohort}
45 patients (143 scans) of the final annotated \kchmcrc data were randomly selected for expert radiological assessment to evaluate model generalisation across multiple data sources. A comparison was then performed between two models: \textbf{Model 1}, trained solely on the heterogeneous \uclhmcrc dataset, and \textbf{Model 2}, trained on a combined dataset that included the \uclhmcrc, as well as the homogeneous \tilburgscgk and \londonscgk gamma knife planning datasets, and the remaining scans from the \kchmcrc dataset.

An expert neuroradiologist (S.C.) reviewed 286 automated segmentations (143 from each model) presented in a randomised format blinded to the segmentation model.
Segmentations were rejected if they met any of the following criteria:
\begin{itemize} 

\item Outlined anatomical structures incorrectly, such as vessels or the cochlea. 

\item Missed discrete portions of the tumour on the same or adjacent slices. 

\item Over-contoured by more than five adjacent voxels on a single axial plane, with the exception of small intracanalicular lesions, where an over-contour that substantially increased the volume was also rejected.

\item Failed to outline focal extensions of the tumour. \end{itemize}

\begin{figure}[ht!]
\centering
\includegraphics[width=0.8\linewidth]{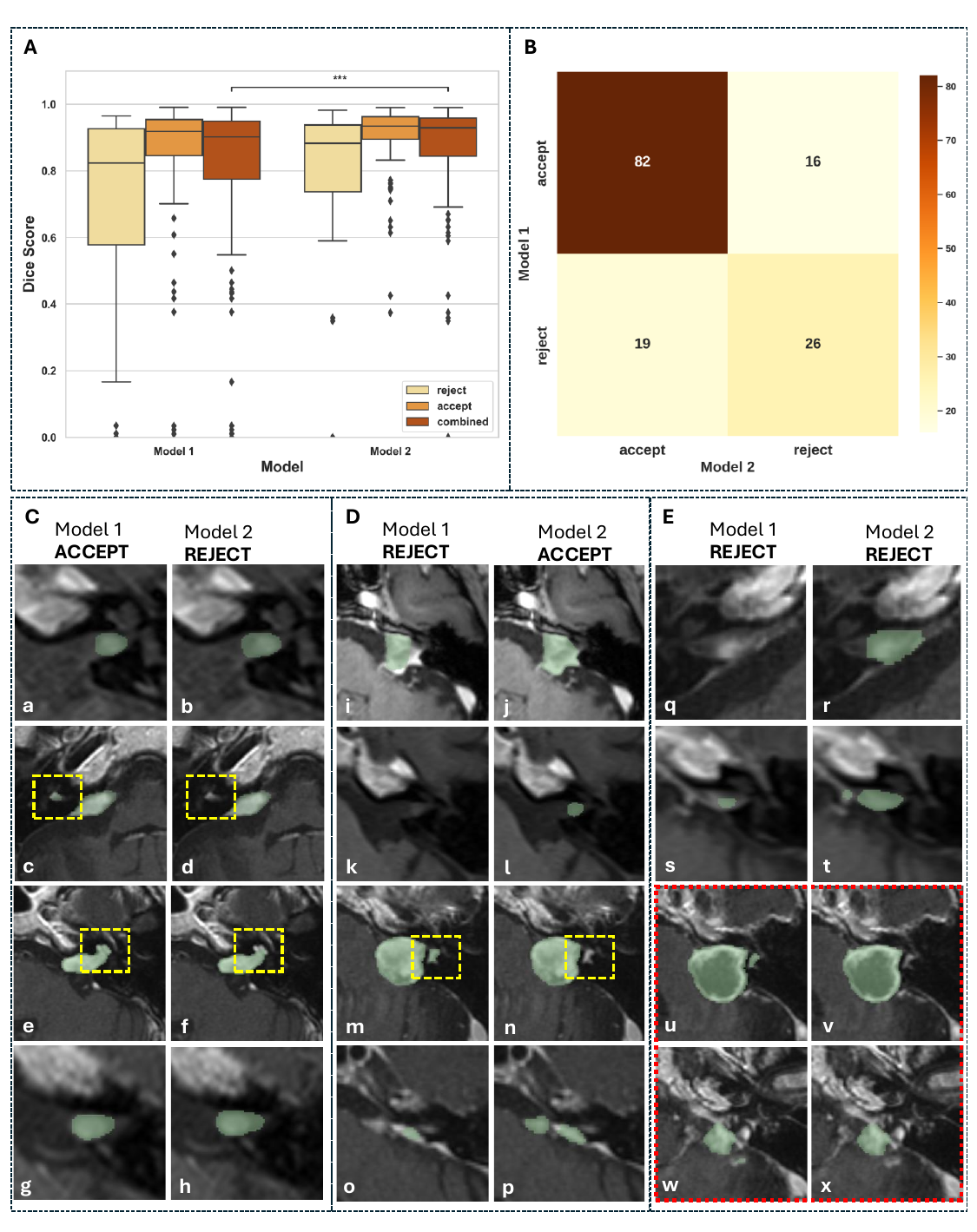}
\caption{(A) Visualisation of the DSC between Model 1 and Model 2 for segmentations grouped by expert decisions of acceptance or rejection. (B) Agreement between Model 1 and Model 2 regarding expert acceptance and rejection decisions. (C)Four example scans where the outcome of Model 1 was accepted, but the outcome of Model 2 was rejected. (D) Four example scans where the outcome of Model 1 was rejected, but the outcome of Model 2 was accepted. (E) Three example scans where both Model 1 and Model 2 outcomes were rejected. The images \textbf{u, v, w, x} are from the same scan, shown on two distinct slices.}
\label{fig:mdmused_comparison}
\end{figure}

The DSC between the two models, with segmentations grouped by expert decision (accept or reject) is visualised in \figref{fig:mdmused_comparison} (A). Model 2 generally shows higher median DSC for both accepted and rejected segmentations, indicating a tendency for better alignment with expert-delineated annotations.
Accepted segmentations have higher DSC with smaller interquartile ranges, reflecting consistent accuracy, while rejected cases display a wider range and lower DSC, particularly in Model 1, which has more outliers. This comparison suggests that Model 2, trained on a more diverse dataset, achieves a slightly higher overall performance and robustness in segmentation accuracy. All accepted and rejected segmentations were combined, and a paired Wilcoxon signed-rank test with Bonferroni correction was conducted to assess the significance of improvements between Model 1 and Model 2, yielding a statistically significant result (p $<$ 0.0001). 

The agreement between Model 1 and Model 2 in terms of expert acceptance and rejection decisions is shown in \figref{fig:mdmused_comparison} (B). Of the total segmentations, 82 were accepted by both models, demonstrating consistent quality in these cases. There were 19 cases accepted by Model 2 but rejected when generated using Model 1, suggesting Model 2's segmentation quality might better meet the expert criteria in certain cases. Conversely, 16 accepted Model 1 segmentations  were rejected when generated using Model 2, indicating specific instances where Model 1 performed better. A total of 26 cases were rejected by both models, highlighting instances where neither model met the required segmentation accuracy.

\figref{fig:mdmused_comparison}(C) and (D) illustrate example cases that were rejected in at least one of the outcomes from Model 1 or Model 2, while both outcomes were rejected in (\figref{fig:mdmused_comparison} (E). \tabref{tab:expert_decisions} lists the reasons for rejections for 12 sample cases as assessed by the expert neuroradiologist. 

\begin{table}[htbp]
\centering
\caption{Decisions for 12 sample scans and reasons}
\begin{tabularx}{\textwidth}{|>{\hsize=.1\hsize}X|>{\hsize=.1\hsize}X|>{\hsize=0.8\hsize}X|}
\hline 
\multicolumn{2}{|c|}{Expert Decision} & \textbf{Explanation} \\ \cline{1-2}
\textbf{Model 1} & \textbf{Model 2} &   \\ \hline
\textbf{a} - accept & \textbf{b} - reject & Model 2 over-contoured a small intracanalicular tumour, significantly increasing the volume, so Model 2 output (b) was rejected. \\ \hline
\textbf{c} - accept & \textbf{d} - reject & A discrete region (outlined in the yellow dashed box on the axial plane) was not detected by Model 2. \\ \hline
\textbf{e} - accept & \textbf{f} - reject & The tumour has extended to the cochlea (highlighted in yellow dashed box), which was not detected by Model 2. \\ \hline
\textbf{g} - accept & \textbf{h} - reject & Model 2 produced an over-segmentation of a focal extension of the tumour, so Model 2 outcome was rejected. \\ \hline
\textbf{i} - reject & \textbf{j} - accept & Model 1 output was under-segmented and therefore rejected. \\ \hline
\textbf{k} - reject & \textbf{l} - accept & Model 1 failed to detect a small intracanalicular tumour, so Model 1 outcome was rejected. \\ \hline
\textbf{m} - reject & \textbf{n} - accept & Model 1 incorrectly segmented a vascular region (indicated by the yellow dashed box). \\ \hline
\textbf{o} - reject & \textbf{p} - accept & In this post-operative case, Model 1 failed to identify the discrete residual tumour. \\ \hline
\textbf{q} - reject & \textbf{r} - reject & Model 1 missed the tumour entirely, while Model 2 over-segmented the tumour. \\ \hline
\textbf{s} - reject & \textbf{t} - reject & Model 1 output was under-segmented, while Model 2 incorrectly segmented a discrete vessel. \\ \hline
\textbf{u} - reject & \textbf{v} - reject & \multirow{2}{\hsize}{ \textbf{u, v, w, x} are from the same scan on two distinct slices. In slice \textbf{u} and \textbf{v}, Model 1 incorrectly segmented a vessel, which was not segmented by Model 2. However, on slice \textbf{w} and \textbf{x}, Model 2 failed to identify a discrete tumour region that was correctly identified by Model 1. Due to failures in distinct slices, both outputs were rejected.} \rule[1ex]{0pt}{2ex} \\ \cline{1-2}
  \textbf{w} - reject & \textbf{x} - reject &   \rule[1ex]{0pt}{3.5ex}  \\ \hline
\end{tabularx} 
\label{tab:expert_decisions}
\end{table}

Interestingly, while the DSC showed a significant increase from Model 1 to Model 2, this improvement did not translate into a corresponding change in expert decisions, which remained consistent across the two models. This suggests that DSC alone may not fully capture the nuances of model performance, particularly in complex cases where segmentation quality requires more than just a high DSC. Such cases likely benefit from additional qualitative assessments or alternative quantitative metrics that better align with expert evaluations, other the DSC, the current gold standard metric for performance assessments of segmentation tasks. This observation highlights the importance of integrating both quantitative scores and expert insights for a more comprehensive understanding of model behaviour and clinical applicability.



\section*{Data availability}
The \kchmcrc dataset is publicly available at \url{https://doi.org/10.7937/bq0z-xa62}. 

\bibliography{sample}

\section*{Author contributions}
N.W. contributed to conceptualisation, methodology, formal analysis, writing, review, editing. M.I, O.M., S.K., S.C., L.G.M. were involved in data curation. A.K., R.B., A.H., S.T., S.B., E.M., R.O., D.J., F.M., K.C., O.A., N.T. were responsible for data collection. T.V. and J.S. provided conceptualisation, supervision, methodology, writing, review and editing.

\section*{Declaration of interests}
T.V. and J.S. are co-founders and shareholders of Hypervision Surgical. We confirm that no products from Hypervision Surgical were used in this study and Hypervision Surgical has no interest in this work. Other authors declare that they have no known competing financial interests or personal relationships that could have appeared to influence the work reported in this paper.

\section*{Funding}
N. Wijethilake was supported by the UK Medical Research Council [MR/N013700/1] and the King’s College London MRC Doctoral Training Partnership in Biomedical Sciences. 
This work was supported by core funding from the Wellcome Trust (203148/Z/16/Z) and EPSRC (NS/A000049/1) and an MRC project grant (MC/PC/180520). 
%
For the purpose of open access, the authors have applied a CC BY public copyright licence to any Author Accepted Manuscript version arising from this submission.

\end{document}